\documentclass[runningheads]{llncs}
\usepackage[T1]{fontenc}
\usepackage{graphicx}
\usepackage{caption}
\usepackage{subcaption}
\usepackage{pgfplots}
\usepackage{svg}
\pgfplotsset{compat=1.18}
\usepackage{float}

\usepackage{hyperref}
\usepackage{url}
\usepackage{textcomp}
\usepackage{todonotes}
\hypersetup{
    colorlinks=true,
    linkcolor=blue,
    filecolor=magenta,
    citecolor=blue,
    urlcolor=blue
}

\begin{document}
\title{BirdRecorder's AI on Sky: Safeguarding birds of prey by detection and classification of tiny objects around wind turbines}
\titlerunning{BirdRecorder's AI on Sky}
%

\author{Nico Klar\inst{1}\orcidID{0009-0006-8308-0318} 
Nizam Gifary\inst{1}\orcidID{0009-0001-8967-3812}
Felix P.G.~Ziegler\inst{1}\orcidID{0000-0002-2485-0398}
Frank Sehnke\inst{1}\orcidID{0000-0002-5742-1805}
Anton Kaifel\inst{1}\orcidID{0000-0002-5818-0193}
Eric Price\inst{2}\orcidID{0000-0002-3480-8054} \and
Aamir Ahmad\inst{2}\orcidID{0000-0002-0727-3031}}

\authorrunning{Klar et al.}
%
\institute{Center for Solar Energy and Hydrogen Research (ZSW), Meitnerstraße 1, 70563 Stuttgart, Germany
\email{nico.klar@zsw-bw.de} 
\and
Flight Robotics and Perception Group, University of Stuttgart, 70569 Stuttgart, Germany
}

\maketitle              

\begin{abstract}
The urgent need for renewable energy expansion, particularly wind power, is hindered by conflicts with wildlife conservation. To address this, we developed \emph{BirdRecorder}, an advanced AI-based anti-collision system to protect endangered birds, especially the red kite  (\textit{Milvus milvus}). Integrating robotics, telemetry, and high-performance AI algorithms, \mbox{BirdRecorder} aims to detect, track, and classify avian species within a range of 800 m to minimize bird-turbine collisions.

BirdRecorder integrates advanced AI methods with optimized hardware and software architectures to enable real-time image processing. Leveraging Single Shot Detector (SSD) \cite{10.1007/978-3-319-46448-0_2} for detection, combined with specialized hardware acceleration and tracking algorithms, our system achieves high detection precision while maintaining the speed necessary for real-time decision-making. By combining these components, BirdRecorder outperforms existing approaches in both accuracy and efficiency.

In this paper, we summarize results on field tests and performance of the BirdRecorder system. By bridging the gap between renewable energy expansion and wildlife conservation, BirdRecorder contributes to a more sustainable coexistence of technology and nature.

\end{abstract}
\section{Introduction}
The current climate crisis and energy dependencies require a massively accelerated expansion of renewable energies worldwide. Wind energy is expected to account for $8.6\%$ of global energy generation in 2024 \cite{windenergy2024}. Particularly in Europe the 
necessary dynamic expansion of wind power stagnates due to conflicts with animal species protection. Versatile AI-based technical avoidance measures provide a key instrument in the near future to achieve a balance between species protection and economic efficiency. The purpose of these anti-collision systems is to detect specific endangered bird species and if necessary to shut or slow down a wind turbine when such a bird is in close proximity to the rotor blades. In this paper, we summarize the status of our anti-collision system BirdRecorder, synergizing state-of-the-art robotics and telemetry with modern high-performance AI methods. 
In 2023, the BirdRecorder system won the AI Champions Award Baden-Württemberg \cite{KI-Champion}. 
Model architectures and algorithms supporting this study are available on GitHub: 
\href{https://github.com/os-simopt/BirdRecorder}{https://github.com/os-simopt/BirdRecorder}.

\subsection{Motivation}
The research project WindForS \cite{clifton2017new}, supported by ZSW, the Universities of Stuttgart, T\"ubingen, TU Munich, the Karlsruhe Institute of Technology and the Universities of Aalen and Esslingen, aims to make wind energy generation more acceptable and cheaper. The goal is to build and operate two research wind turbines in the complex terrain of the Swabian Jura, including measuring stations at four locations.
In order to ensure compliance with species protection regulations and to guarantee continuous test operation throughout the year, the accompanying nature conservation research projects NatForWINSENT \cite{bfn_natforwinsent2}, \cite{musiol2023umsetzung} and BirdRecorder \cite{kaifel2023birdrecorder}, funded by the German Federal Agency for Nature Conservation (BfN), were launched at the same time in 2019. 
The aim was to develop a prototype of an anti-collision system to minimize bird strikes at wind turbines. In the Swabian Jura region, research concentrated on the detection of the red kite (\textit{Milvus milvus}). The main focus was on the detection, tracking and subsequent species classification of flying animals at a range of up to 800 m to ensure timely turbine shutdown. 
\subsection{State-of-the-Art}
The identification of avian species in flight is primarily used for habitat analysis, natural pest control in agriculture, collision avoidance at airports and in the wind energy industry.
In the past, ornithologists manually conducted these observations, spending days recording flight paths using laser range finders (LRFs).
Nowadays, airports primarily rely on automated systems. Here, radar-based systems are commonly used to detect birds near wind turbines, as the goal is to determine the size of the bird regardless of its species. This method is advantageous because it is not affected by weather conditions such as fog, rain, or direct sunlight \cite{RobinRadar}. On the other hand, it is important to differentiate between bird species at risk from wind turbines and to shut down the turbines when necessary. Currently, the most common method of doing this is visual identification using camera images. As of 2024, only one system in the world, IdentiFlight \cite{Identiflight}, has demonstrated the detection range of 500--700\,m \cite{KNE} required for timely shutdown. However, publicly available documentation on comparable bird detection systems is limited, and standardized testing or acceptance procedures are lacking both in the literature and in practice, making direct comparisons challenging.

\section{Methods and setup}
\subsection{Hardware}
\label{sec:hardware-setup}
\begin{figure}
\centering
\includegraphics[width=0.53\textwidth, angle=0]{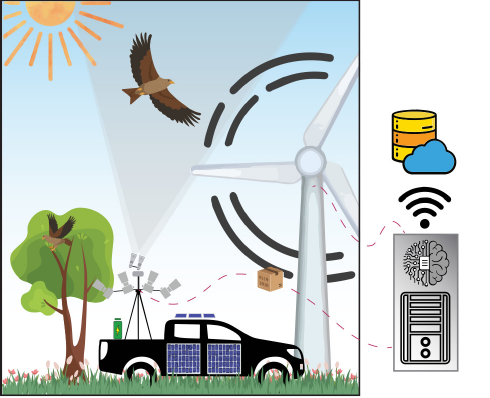}
\caption{Sketch of the BirdRecorder mobile sensor rig featuring static cameras and a movable stereo pair. Data is transmitted to a processing unit that performs real-time analysis and turbine shutdown in case of red kite collision risk. Additionally, all detected objects and related data are archived in a cloud database.}
\label{fig:system-sketch}
\end{figure}

The system consists of two identical sensor rigs, each covering half of the sky. Each half of the system contains four statically mounted cameras, along with a centrally positioned pan-tilt-unit (PTU) hosting two cameras with tele objectives in stereo configuration for distance estimation. The static cameras have shorter focal lengths, allowing for wide-area sky scanning to detect regions of interest. Conversely, the tele-cameras have longer focal lengths, enabling higher resolution focusing on specific areas of interest within the sky. A sketch of the sensor system is shown in Figure~\ref{fig:system-sketch}.

\vspace{-0.8em}
Each half of the system is controlled by an industrial PC. 
Communication between all components is facilitated via Ethernet. 
The cameras are linked to the industrial PC via coaxial 
cables connected to a frame grabber. The captured image streams are processed in real time at 4\,Hz using four NVIDIA RTX~4070 GPUs, enabling full 360\textdegree{} coverage.
The current hardware setup offers a practical trade-off between detection performance and computational cost, but further optimization may be needed for large-scale deployments.

\subsection{Software}
\begin{figure}
\centering
\includegraphics[width=1\textwidth, angle=0]{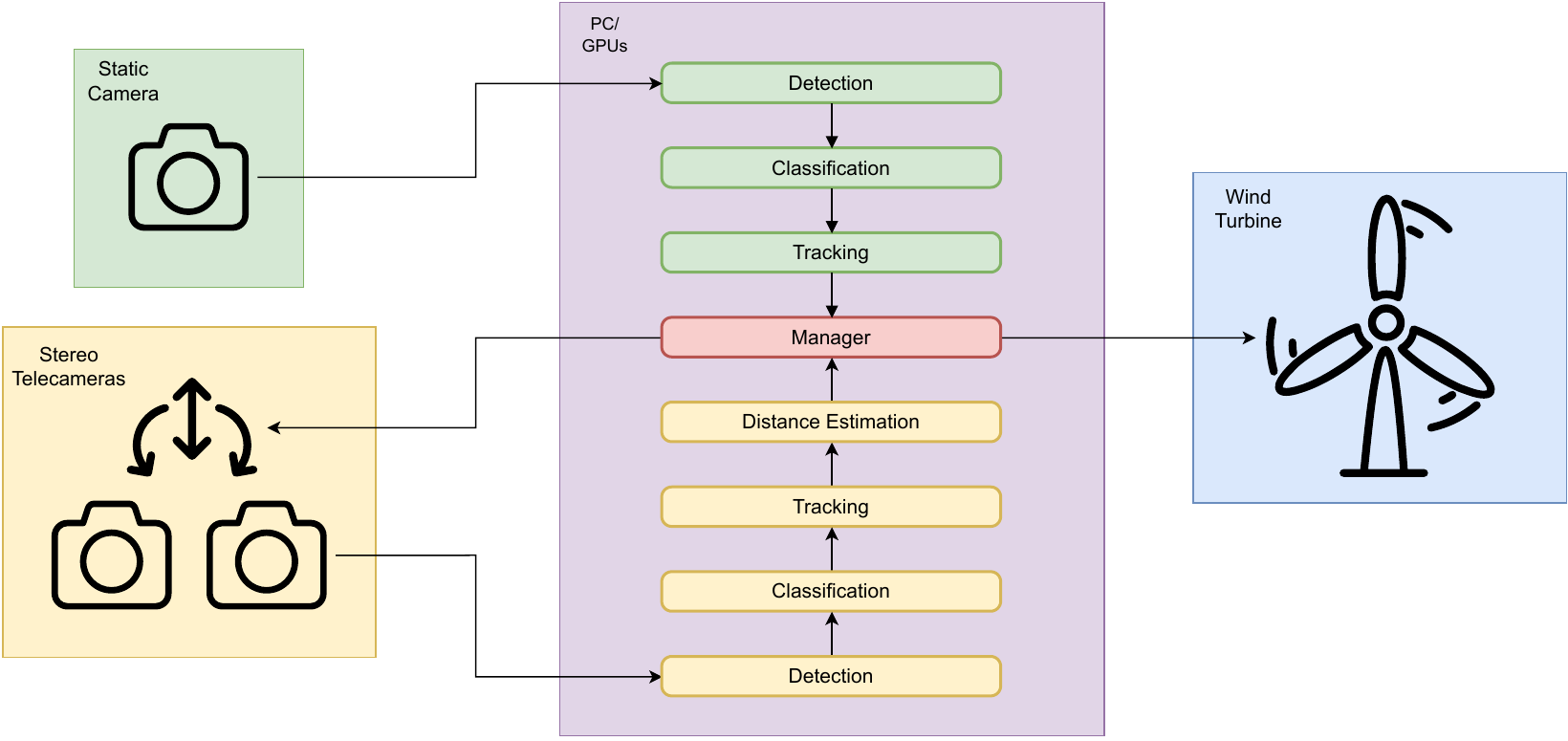}
\caption{Overview of the hard and software pipeline. Data streams coming in from statically mounted cameras (green) are processed to detect, classify, and track endangered bird species sequentially for each camera frame, ensuring continuous real-time processing. A central priority management unit (red) handles the decision making
and controls the movable tele cameras (yellow) to obtain a full 3D track of the object (bird) of interest. If the protected bird flies into
the danger zone, the wind turbine will shut down.}
\label{fig:system-pipeline}
\end{figure}
The software architecture relies on ROS2 for facilitating data exchange within the program. ROS2 is organized around the concept of nodes, which are modular components tailored for specific tasks. These nodes communicate in a publish-subscribe manner, wherein they can both send and receive data through topics~\cite{Macenski_2022}. 

The static cameras capture raw images continuously, covering the entire 360-degree field of view. The detection node processes these images to identify regions of interest (ROIs) where avian species can be found. The classification node then analyses these ROIs to assign object labels. After classification, the ROIs are sent to the tracking node, which assigns unique identifiers to the detected objects and monitors their movement across successive frames, using a variation of the Kalman filter. The manager node then receives this information, selects the object for tracking, and directs the PTU accordingly. The same process is repeated for tele-cameras mounted on the PTU, but focusing on a single object determined by the manager node. 
The system estimates distance via stereo‐triangulation:\\[-0.8\baselineskip]
\[
  Z = \frac{f\,B}{\delta}
\]\\[-0.8\baselineskip]
with \(Z\) the camera–object distance, \(f\) the focal length, \(B\) the baseline (camera‐center spacing), and \(\delta\) the disparity (pixel shift).
This data is then transmitted to the manager node. If the object identified belongs to a protected species and is within a designated danger zone, the manager sends a command to stop turbine operations. An overview of the software pipeline is displayed in Figure~\ref{fig:system-pipeline}.

\subsection{AI methods}\label{subsec:ai_methods}
The AI methods serve as the central hub of the overall system. In this section, images are converted into information that is used by the anti-collision system to make further decisions.
Two cutting-edge technologies are employed: object detection and object classification.

\subsubsection{Detection:} The Single Shot MultiBox Detector (here SSD300 version) was selected due to its high speed and efficiency. This AI model is known for its higher accuracy in detecting smaller objects compared to other state-of-the-art models like You Only Look Once (YOLO) \cite{redmon2016look}. This is due to SSD's use of multiple feature maps at different scales to detect objects of various sizes. By incorporating feature maps from different levels of abstraction, SSD can effectively capture and represent objects with finer details, including smaller objects. This multi-scale approach allows SSD to better handle variations in object sizes within an image. This is especially relevant for this use case because objects of interest, for example endangered birds, flying at 800 m are only represented by a few pixels in the images whereas at 300 m they correspond to over a hundred pixels. 
SSD is also well known for its very precise localization of detected objects, which is a prerequisite for aligning the tele cameras to the correct position in the sky.

\subsubsection{Classification:} 
\begin{figure}[htbp]
\includegraphics[width=1.0\textwidth, angle=0]{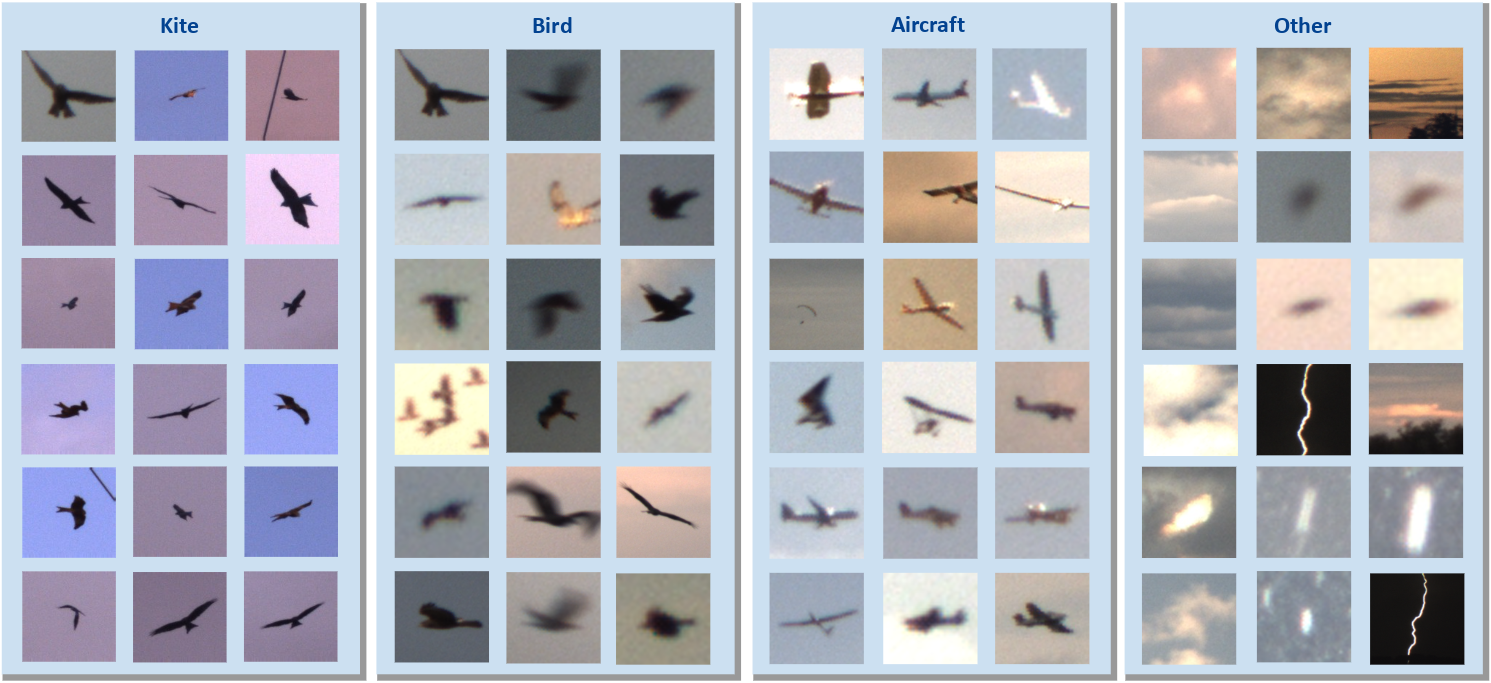}
\caption{This image shows example ROIs for the four selected classes.}
\label{fig:classes}
\end{figure}

Since the problem of classifying tiny objects is a very challenging task, finding a suitable model for our needs proved difficult. Although SSD inherently performs both detection and classification, multi-task models like SSD tend to underperform in classification when training data is limited. Given the high cost and effort required for manual annotation of tiny aerial objects, our dataset is constrained. In our cascaded, two-step approach, the SSD’s classification is used only to filter out obvious outliers, while a dedicated single-task CNN—designed with convolutional layers for feature extraction, pooling layers for spatial downsampling and translation invariance, a flattening layer for vectorizing the feature maps, fully connected layers for high-level representation learning, and a softmax output layer for multi-class classification—delivers a refined classification. Four classes were defined for prototype development: Kite, Bird, Aircraft, and Other (capturing artifacts such as clouds, lightning, or reflections), as shown in Figure~\ref{fig:classes}. This approach significantly enhances system robustness and can be easily extended to include additional bird species.

\section{Data}
Data serves as the foundation for AI algorithms. Due to the unique use case of identifying tiny birds, there were no publicly available datasets to use at the beginning of the project. As a result, related datasets for bird detection in flight such as \cite{Kondo_2023} and \cite{7351607} were just utilized for pre-training. 
Over the course of several measurement campaigns from 2019 to 2021, 18,792,026 images were recorded in various weather conditions and seasons.

\begin{table}
\centering
\caption{This table displays the number and size of the recorded images.}\label{tab_data}
\begin{tabular}{|p{1.5cm}|p{3cm}|p{3cm}|}
\hline
Year & Number of Images & Amount of Data\\
\hline
2019 & 3,979,890 & 30.36 TB\\
2020 & 12,024,807 & 91.74 TB\\
2021 & 2,787,329 & 21.27 TB\\
\hline
\end{tabular}
\end{table}    

To filter out irrelevant images, motion detection was performed using frame differencing, as not all images contain flying objects. The algorithm developed identifies motion in the sky by detecting pixel differences between consecutive frames in image sequences. After applying DBSCAN to these pixels, continuously moving objects, such as neighboring wind turbines or tree tops, were masked. The method identified and cropped 23,861,099 objects/ROIs, each measuring 128x128 pixels.

\subsection{Pre-annotation using autoencoder} \label{subsec:autoencoder}
\begin{figure}[htbp]
\includegraphics[width=1.0\textwidth, angle=0]{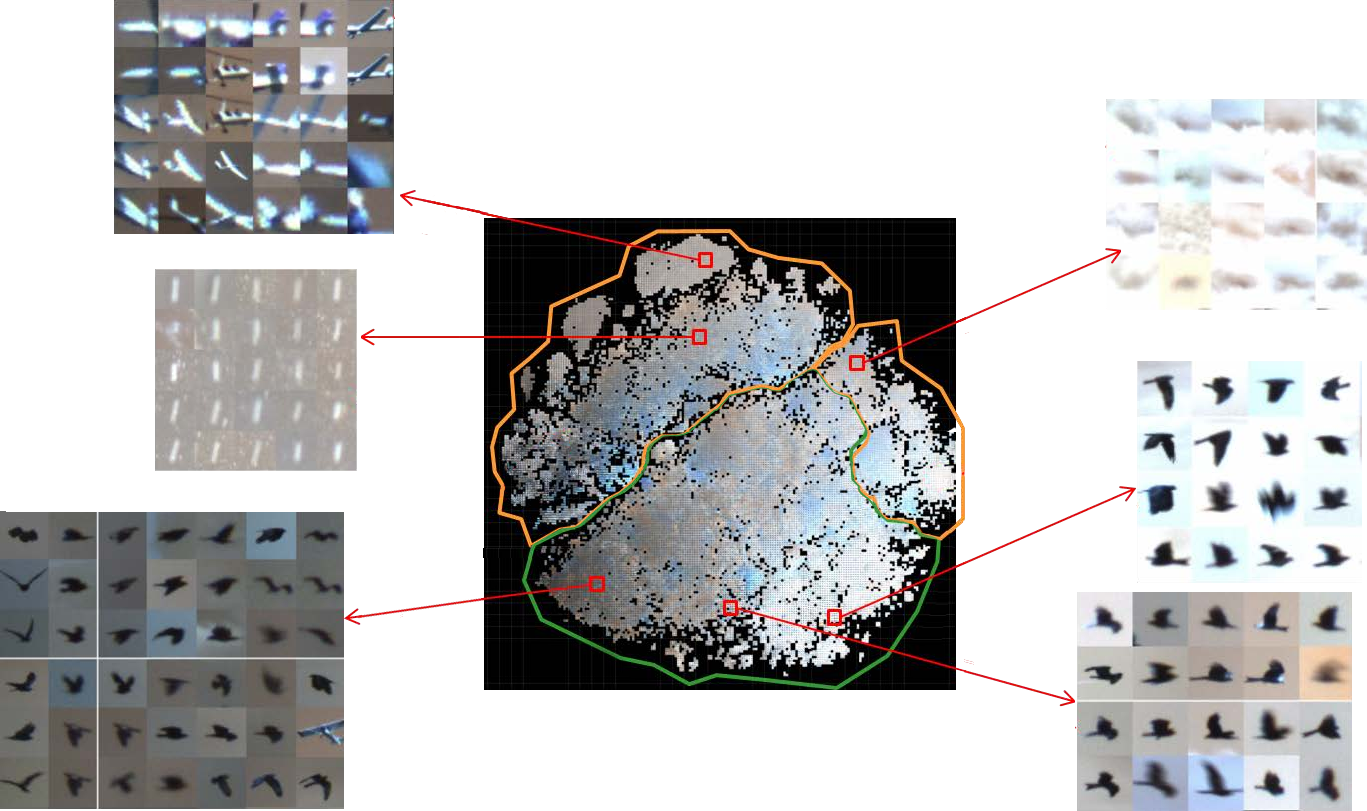}
\caption{Illustration of our method for unsupervised clustering the training data. Low-dimensional vectors obtained from an autoencoder are clustered using t-SNE. Thumbnails showcase representative samples from bird and non-bird clusters, demonstrating the effectiveness of our approach for image pre-classification.}
\label{fig:autoencoder}
\end{figure}

AI-driven clustering methods were developed to provide ornithologists with a representative set of ROIs, that capture a large range of different birds while excluding irrelevant objects. An autoencoder was used to learn a lower-dimensional representation of the images. The encoder part of the autoencoder extracts features (represented as a one-dimensional vector of length 1024, technically known as a latent representation), which were then clustered using t-Distributed Stochastic Neighbor Embedding (t-SNE) \cite{JMLR:v9:vandermaaten08a}. Next, a representative subset of images was selected from various clusters based on specific criteria, see Figure~\ref{fig:autoencoder}. Finally, the chosen subset was evaluated for quality and effectiveness in capturing the diversity of the original image dataset.

\subsection{Annotations by ornithologists}
A subset of 680,000 ROIs was selected using the autoencoder and t-SNE presented in section \ref{subsec:autoencoder}. Several experts in the field of ornithology then manually annotated these ROIs. To aid in this process, we developed our own custom labeling tool connected to a database, specifically tailored to our application. This tool allowed entire flight sequences to be labeled in one step. This approach provided two advantages:
\begin{itemize}
    \item[(1)] Identifying a bird from a single image section can be challenging. Therefore, experts prefer using image sequences for a more accurate assessment.
    \item[(2)] Direct labeling of the entire sequence enables annotation of a larger number of images in the same amount of time.
\end{itemize}

\subsection{LRF recordings and drone data}
LRF measurements were taken repeatedly during the breeding season as part of the \mbox{NatForWINSET II} project to study bird behavior near wind turbines. The combination of these measurements with recordings from the BirdRecorder system provided valuable data for both classification and distance estimation. Despite the high cost and time required for LRF measurements, they are a reliable means of evaluating the system.
In addition, three drones could be used, each positioned to an accuracy of two centimeters using differential GPS. This data can also be used to assess the algorithms. In addition, certain scenarios can be reproduced using stored flight routes. 
\subsection{GPS tracking}
In 2019 and 2020, we attempted to equip red kites from an eyrie near the WINSENT wind turbines with GPS trackers. 
Two red kites were successfully tagged. Their positions were obtained within the geofence around the wind turbine with a resolution of up to 0.2 Hz. It is important to note that GPS positions may be inaccurate on the horizontal x-y plane and especially with regard to the vertical z direction (height) \cite{10.1371/journal.pone.0199617,LiuGPS,Poessel,Peron}. Nevertheless, the GPS data is helpful in enhancing and verifying classifications.

\subsection{Test dataset}
In our research, we introduced an independent test dataset encompassing a wide range of scenarios and seasons, including variations in daylight brightness, diverse weather conditions, and different angles of sunlight. The dataset comprises more than 100{,}000 labeled images, corresponding to 15\% of our total annotated data. It was designed to provide a balanced representation of the four target classes: \textit{kite}, \textit{bird}, \textit{aircraft}, and \textit{other}, each comprising 25\% of the test samples.

\begin{figure}[htbp]
\includegraphics[width=1.0\textwidth, angle=0]{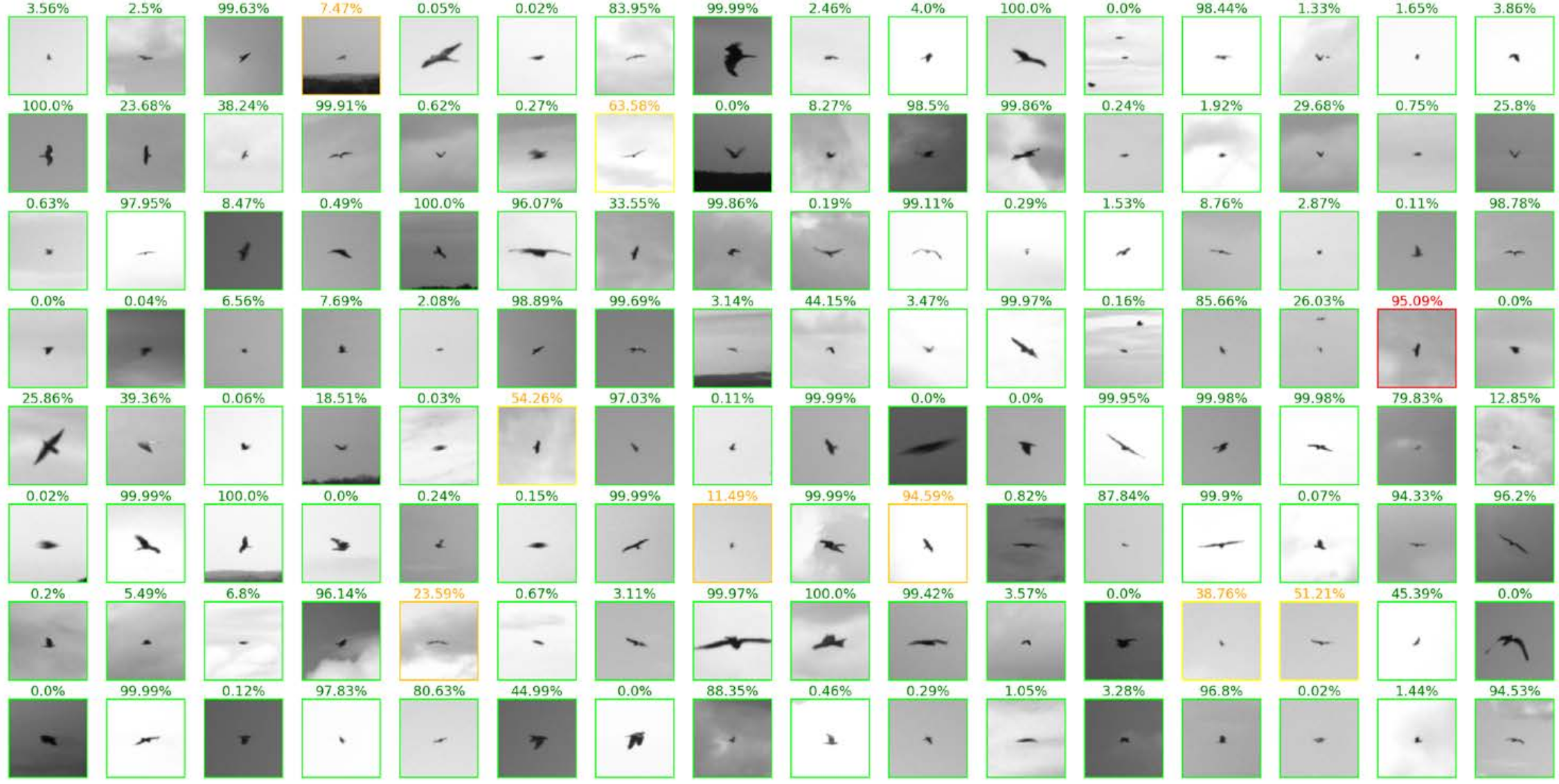}
\caption{AI-driven methods can effectively detect possible anomalies in the test data set and provide corrections to ornithologists. Through several iterations, a highly accurate and reliable test dataset was created. Green-bordered images were correctly classified, while yellow, orange, and red-bordered images were misclassified. Border color indicates the network's confidence in misclassification, ranging from yellow (up to 75\% confidence), to orange (85--95\% confidence), to red (over 95\% confidence).}
\label{fig:test_dataset}
\end{figure}

The selection process involved a rigorous multi-stage pipeline combining classical image processing with AI-based anomaly detection, followed by thorough manual validation, see Figure~\ref{fig:test_dataset}. This ensured the creation of a highly accurate and representative test dataset, serving as a robust benchmark for evaluating the performance of our AI models under diverse real-world conditions.

\section{AI-driven real-time image processing}
\subsubsection{The challenge}

Developing a robust and generally deployable anti-collision system in various complex terrains and
weather scenarios demands mastering a few challenging computing tasks:

\begin{itemize}
    \item[(1)] Input data streams from several cameras (ca. 25 MB per image) have to be processed at rates
    of 4Hz and higher.
    \item[(2)] Objects of interest, e.g. endangered birds must be detected over a range up to 800 m across the full sky. In particular, this requires a performant detection of tiny objects.
    \item[(3)] The objects must be classified in real-time.
    \item[(4)] Accurate tracks of the 3D trajectory for full flight path information have to be generated.
    \item[(5)] Following the objects with the movable cameras along their flight path requires 
    performant, highly precise cybernetics.
\end{itemize}

\begin{figure}[htp]
\centering
\begin{subfigure}{0.495\textwidth}
  \centering
  \includegraphics[width=\linewidth]{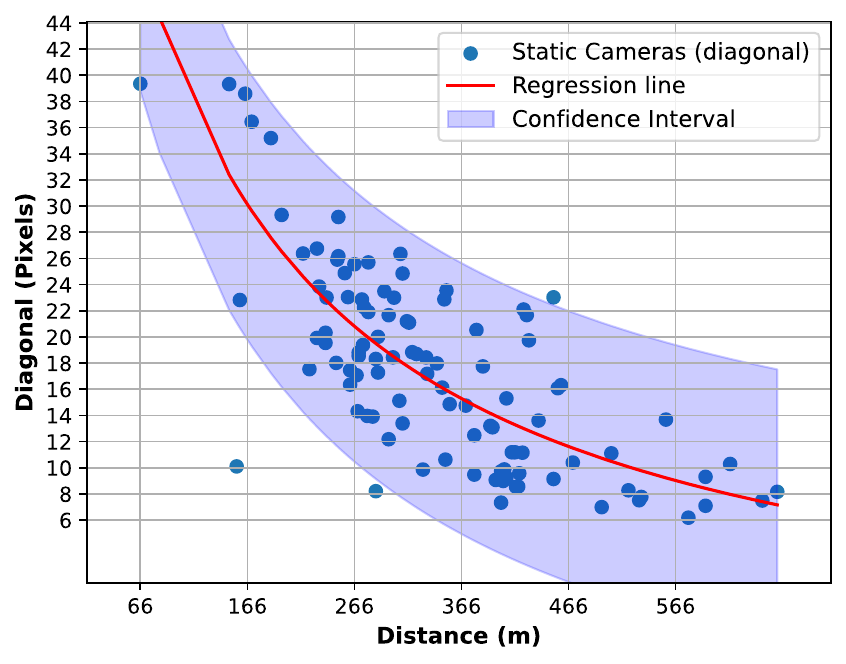}
  \caption{Static Cameras}
  \label{subfig:distance_static}
\end{subfigure}\hfill
\begin{subfigure}{0.495\textwidth}
  \centering
  \includegraphics[width=\linewidth]{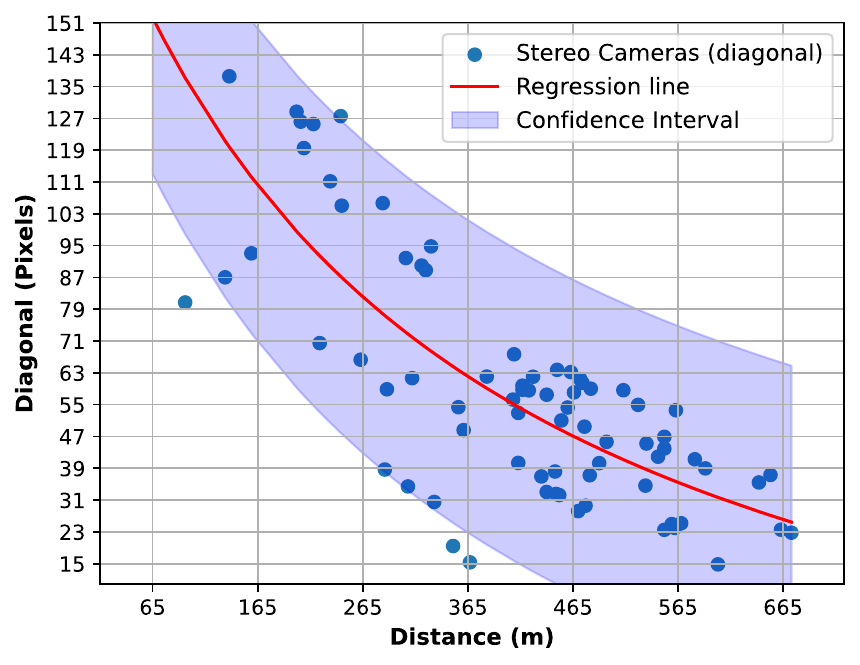}
  \caption{Telemetric Stereo Cameras}
  \label{subfig:distance_stereo}
\end{subfigure}
\caption{Scatter plots of bounding box diagonal (pixels) versus red kite distance (m) for (a) Static Cameras and (b) Stereo Cameras. A nonlinear regression ($\frac{a}{x+b}+c$) with a 95\% confidence interval (shaded) illustrates the inverse relationship between distance and apparent object size.}
\label{fig:distance}
\end{figure}

The following sections focus on object recognition. A preliminary analysis of the relevant objects was necessary.
To determine the blob sizes of interest, we intersect LRF measurements of red kites with camera images captured. Figure~\ref{fig:distance} shows the ratio of distance to the number of pixels captured in a diagonal bounding box. This regression yields a calibration curve predicting distance by blob size, crucial for tuning detection thresholds and enhancing accuracy.

\subsection{Traditional image processing approaches}\label{subsec:traditionalproc}
In previous stages of the BirdRecorder project, the entire image acquisition and processing pipeline was implemented on field-programmable gate arrays \mbox{(FPGAs)} \cite{fpga}, which are well known for their real-time performance e.g., for autonomous driving assistance systems.
Traditional image processing algorithms such as detecting motion 
via difference images in connection with a blob detection 
for finding ROIs were used. 
Adaptability to different weather scenarios
was implemented via a filter to discriminate airborne objects and avian species from clouds.
To avoid false positive detections, we had to mask out constantly moving objects such as treetops and wind turbines. Unfortunately, this resulted in blind spots that must be excluded from detection.
Another disadvantage of the traditional methods lies in low dimensional parametrizations 
(10 parameters) coming with the need for laborious tuning depending on the weather condition.
Despite being fast FPGAs come with the
disadvantage of exhaustive low-level programming even for simple tasks. In contrast,
GPUs offer both high performance and come with a plethora of libraries and SDKs for image processing, offering an accessible implementation and deployment of AI methods in particular. 
For the use case at hand implementing state-of-the-art AI algorithms on GPUs 
overcomes the limitations of using traditional algorithms on FPGAs.
At present we restrict the use of FPGAs to image acquisition from CXP cameras 
allowing for a high frequency of the incoming data stream. 

\subsection{AI-driven detection}
As described in \ref{subsec:traditionalproc}, the arduous task of parameter tuning for adaptation presents a significant challenge in addressing all scenarios.
However, an even greater disadvantage is the inability to recognize objects in front of moving areas. This results in the repeated loss of tracked flying objects and the inability to continuously record trajectories.
AI enables us to overcome these issues and surpass all of the classical image processing methods we previously employed (Section \ref{subsec:traditionalproc}).

For this purpose, we implemented the Single Shot MultiBox Detector with a receptive field of 300x300 pixels. We used NVIDIA Data Loading Library (DALI) for data preprocessing and augmentation of brightness, contrast, size, shear, and horizontal flips.
We optimized the following parameters and approaches in an extensive series of experiments:
 \begin{itemize}
     \item Optimization of batch size
     \item Selection of the SSD backbone for feature extraction (MobileNetV2 \cite{sandler2018mobilenetv2}, ResNet18, 34, 50, 101, 152 \cite{He2015})
     \item The benefits of pre-training using ImageNet \cite{deng2009imagenet}, SOD4SB \cite{Kondo_2023}, and/or the dataset described in \cite{7351607}
     \item Balancing the class occurrence as well as object sizes
     \item Using learning rate warm-up and/or decay
 \end{itemize}
The stopping point was determined by early stopping due to the progression of the average precision (AP) score on the validation split.

\subsubsection{Slicing Aided Hyper Inference}
Given that our images have a resolution of 5328x4608 pixels, it is not possible to compress them to the necessary 300x300 pixels. The significant loss of details will greatly affect the detection performance, due to the limitations inherent in its pixel count and level of detail. The Slicing Aided Hyper Inference (SAHI) \cite{9897990}  mitigates these limitations by preserving high-resolution raw images. SAHI offers a novel solution to the challenges posed by traditional sliding window approaches in object detection. This approach segments images into overlapping patches and processes each patch independently, significantly reducing computational overhead while maintaining high detection accuracy. The use of this technique is advantageous for detecting tiny objects because it enables a more detailed analysis, reducing problems related to object occlusion and scale variations that are often encountered in traditional sliding window techniques. Consequently, it exhibits superior performance in detecting tiny objects across various datasets, demonstrating its potential for improving object detection systems in real-world applications.

\subsubsection{Final Assessment: The SSD \& SAHI Advantage}
Popular object detectors like YOLO and Faster R-CNN often struggle with extremely small objects in high-resolution images. In contrast, our SSD-based approach leverages multi-scale feature maps and precise localization to capture distant, tiny objects with high inference speed, crucial for real-time applications. Moreover, integrating SAHI (Slicing Aided Hyper Inference) enhances performance by processing overlapping image patches, mitigating resolution loss and improving detection. Together, SSD and SAHI deliver an optimal balance of speed and precision—exactly what is essential for our application.

\begin{figure}[htp]
\centering
\begin{subfigure}{0.48\textwidth}
  \centering
  \includegraphics[width=\linewidth]{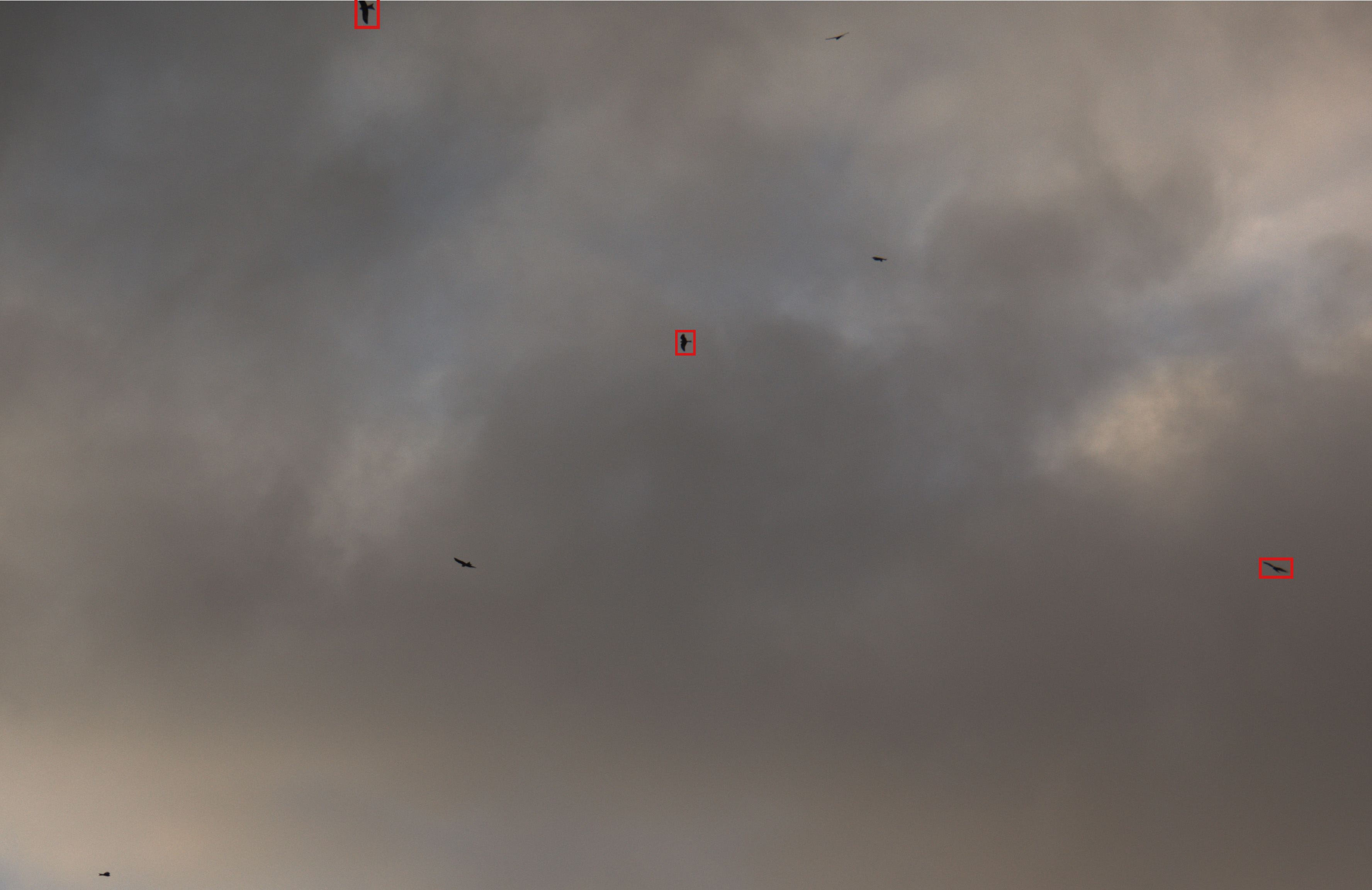}
  \caption{Using Conventional Sliding Window}
  \label{subfig:without-sahi}
\end{subfigure}%
\hspace{0.02\textwidth} 
\begin{subfigure}{0.48\textwidth}
  \centering
  \includegraphics[width=\linewidth]{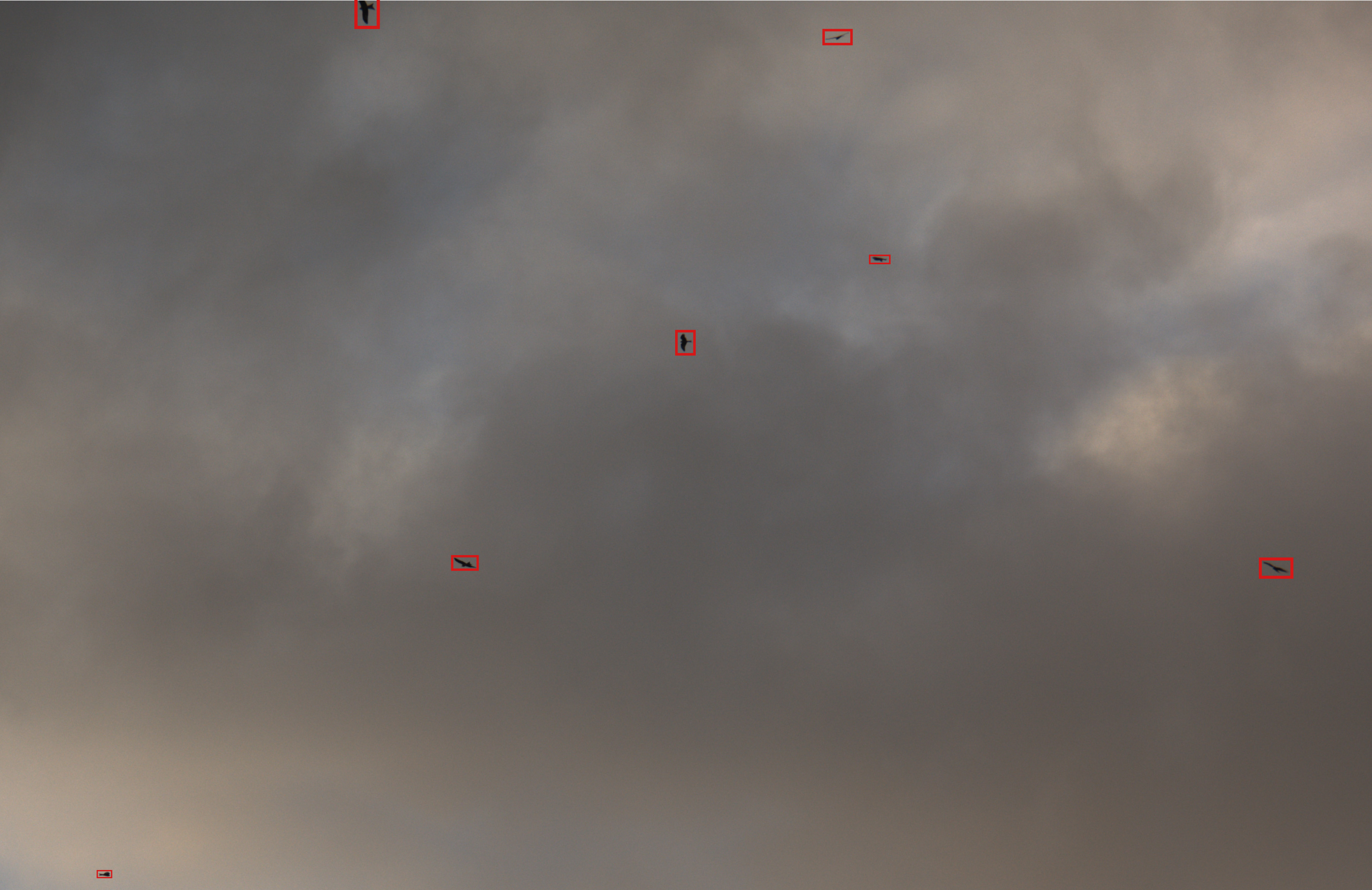}
  \caption{Using SAHI Approach}
  \label{subfig:with-sahi}
\end{subfigure}
\caption{Presented is a magnified view of our detection. The use of Slicing Aided Hyper Inference (SAHI) shows that much more birds can be detected using this method.}
\end{figure}

\subsection{AI-driven classification}
As described in section \ref{subsec:ai_methods}, we optimized a neural network architecture specialized in classifying tiny objects. In our application, we focused on distinguishing the red kite from other bird species, such as the buzzard, which is not part of the endangered species.
Since the feature variance in the classification problem at hand is rather low on the one hand, the visibility of the related objects can be quite bad on the other hand.
This and the number of training images is in stark contrast to datasets like ImageNet or CIFAR, there the number of images is three orders of magnitude higher and visibility of the objects is very good, but also the feature variance is much higher.
As a consequence none of the models currently used for these standard benchmark datasets worked well on our data.
We therefore settled on a pure CNN with 7 convolutional layers and 128 feature maps (FM) in the first layer.
Every convolutional layer halves the resolution by striding and doubles the number of FMs, effectively halving the number of units from layer to layer.
Finally, the output is passed to a dense layer and then a softmax layer.
A single dropout layer is used between the last convolution- and dense layer for regularisation.
This architecture is sometimes called classical autoencoder architecture, but with striding instead of max pooling for resolution reduction.
Nearly all hyper-parameters of this architecture were optimized, partly by hand, partly by using Policy Gradients with Parameter-based Exploration (PGPE~\cite{sehnke2010parameter}), using the validation set performance as reward signal.
For the classification results see Section~\ref{sec:oper-results}.

As distinguishing bird species at the needed distances is an even more challenging task than detection, we closely collaborated with ornithologists to ensure accuracy. We examined the training dataset for anomalies using our AI model's predictions in several iterations. Any anomalies were then passed on to the experts for a final correction decision, similar to our test dataset, Figure~\ref{fig:test_dataset}.

\begin{figure}[htp]
\centering
\begin{subfigure}[b]{0.35\textwidth}
  \centering
  \includegraphics[width=\linewidth]{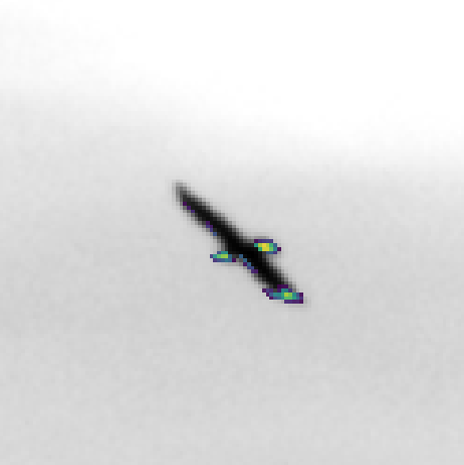}
  \captionsetup{width=1.3\linewidth}
  \caption{Classified as buzzard: Due to the poorly recognizable pose of the red kite, the focus is on the head and tail. However, the characteristic tail is not clearly identifiable as a forked tail. The wing shape is also not typically curved for a red kite from this angle of view. Therefore there is no focus on it.}
\end{subfigure}
\hspace{0.15\textwidth}
\begin{subfigure}[b]{0.35\textwidth}
  \centering
  \includegraphics[width=\linewidth]{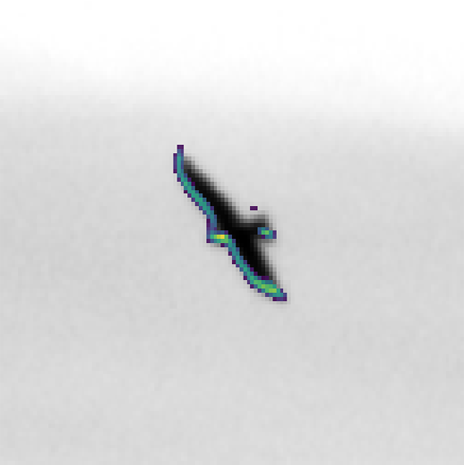}
  \captionsetup{width=1.3\linewidth}
  \caption{Classified as \textit{kite}: Only 300 ms later, the red kite pose becomes clearly recognizable. Grad-CAM provides direct insight into the AI model's focus on the typical wing shape and the tail of the bird.\\\\}
\end{subfigure}
\caption{Visualisation of Grad-CAM using the direct example of a red kite flight.}
\label{fig:grad_cam}
\end{figure}

In the data collection and training optimization phase, it was essential to gain insight into the specific aspects or characteristics where our AI models focus on, especially when dealing with visually similar species. The increasing emphasis on explainable AI highlights the necessity of understanding the decision-making processes of these models, particularly in domains where transparency is crucial.
To demonstrate the effectiveness of explainable AI methods, we used several techniques, including Gradient-weighted Class Activation Mapping (Grad-CAM) \cite{selvaraju2017grad}. Grad-CAM combines the interpretability of generalized additive models with the explanatory power of gradient-based techniques, providing insights into the importance of different features for model predictions.
To demonstrate the effectiveness of Grad-CAM, we provide an example in Figure~\ref{fig:grad_cam}, where the method highlights the essential features used by our AI model to differentiate between visually similar recordings.

\section{Operational results}\label{sec:oper-results}

In this section we summarise various results and benchmarks addressing both technical 
specifications of the BirdRecorder system as well as performance and operations in the field.

\subsection{Performance of the SSD}
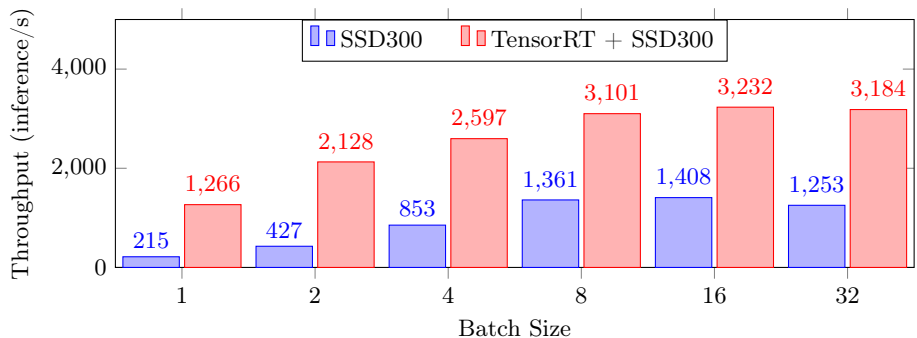
\begin{figure}
    \centering
    \begin{tikzpicture}
    \begin{axis}[
            ybar,
            bar width=.75cm,
            width=1\textwidth, 
            height=0.4\textwidth, 
            legend style={at={(0.5,1)},
                anchor=north,legend columns=-1},
            symbolic x coords={1, 2, 4, 8, 16, 32},
            xtick=data,
            nodes near coords,
            nodes near coords align={vertical},
            ymin=0,ymax=5000,
            xlabel={Batch Size},
            ylabel={Throughput (inference/s)},
        ]
        \addplot table[x=batch_size,y=benchA]{\mydata};
        \addplot table[x=batch_size,y=benchB]{\mydata};
        \legend{SSD300~~~~~, TensorRT + SSD300}
    \end{axis}
\end{tikzpicture}
    \caption{Inference results with TensorRT in units of 300 x 300 sized images per second. The measurements were carried out on an Nvidia RTX 4070 GPU.}
    \label{fig:tensorrt}
\end{figure}

In our system, we apply NVIDIA TensorRT \cite{tensorrt} to optimize the Single Shot MultiBox Detector (SSD) neural network architecture — using layer fusion, kernel selection, and quantization. This yields an approximately threefold increase in throughput over the non‐optimized network (Fig.~\ref{fig:tensorrt}). Benchmarks on an NVIDIA RTX~4070 GPU show that TensorRT+SSD300 reaches a peak throughput of 3232\,inference/s (batch size~16), corresponding to 0.31\,ms per $300\times300$ frame.

\pgfplotstableread[row sep=\\,col sep=&]{
    batch_size & benchA & benchB \\
    1          & 215    & 1266   \\
    2          & 427    & 2128   \\
    4          & 853    & 2597   \\
    8          & 1361   & 3101   \\
    16         & 1408   & 3232   \\
    32         & 1253   & 3184   \\
    }\mydata
     
 \subsection{BirdRecorder validation test}
 The BirdRecorder system was validated with respect to both detection
 and classification rates of red kites. A description of the sensor rig consisting 
 of static cameras for object detection and movable cameras for stereo vision and 3D flight path tracking 
 is described in Section~\ref{sec:hardware-setup}
 A baseline was provided by laser range finder (LRF) measurements
 conducted by a team of ornithologists from Suisse Ornithological Institute (Sempach, Switzerland)
 in fall 2022.
 German policy recommendations suggest that anti-collision systems are to detect endangered species up
 to 700 m in order to ensure sufficient reaction time for slowing down wind turbines \cite{KNE}.
 For the BirdRecorder the 2022 test yielded the following results.

 \textbf{(i) Detection rate:} In a distance range of 0-350 m the detection rate amounted to 85\% of the measured LRF tracks of red kites. In the range 351-700 m, 50\% of the flight tracks were detected. In the latter case the lower performance is strongly correlated 
 with the ill-conditionedness of the stereo vision problem. For a related study 
 see \cite{10.1007/978-3-319-10605-2_7}.
 
 \textbf{(ii) Classification rate:} 
 The BirdRecorder system was able to correctly classify 
 98 \% of the red kites tracked by LRF measurements.
 Note that for this preliminary test, the BirdRecorder system used the naive blob detection
 described above in Section~\ref{subsec:traditionalproc}. Future versions of the system make use of 
 state-of-the-art AI techniques. It is to be expected that detection rates will improve in future
 system versions. 

 Weather conditions also impact detection and classification rates. Often, this influence is not attributable to a single factor but results from the interplay of multiple factors, such as heavy cloud cover and sun reflections. In operational mode, internal logic compensates for these false detections.

\subsection{ML results}
After optimising our ML methods, we applied them to the test datasets corresponding to short (static cameras) and long (tele cameras) focal lengths with respect to detection as well as classification. The results are presented separately for SSD detection and classification below.

\subsubsection{Performance of SSD compared to traditional blob detection}
\begin{figure}[H]
\centering
\begin{subfigure}{0.56\textwidth}
  \centering
  \includegraphics[width=\linewidth]{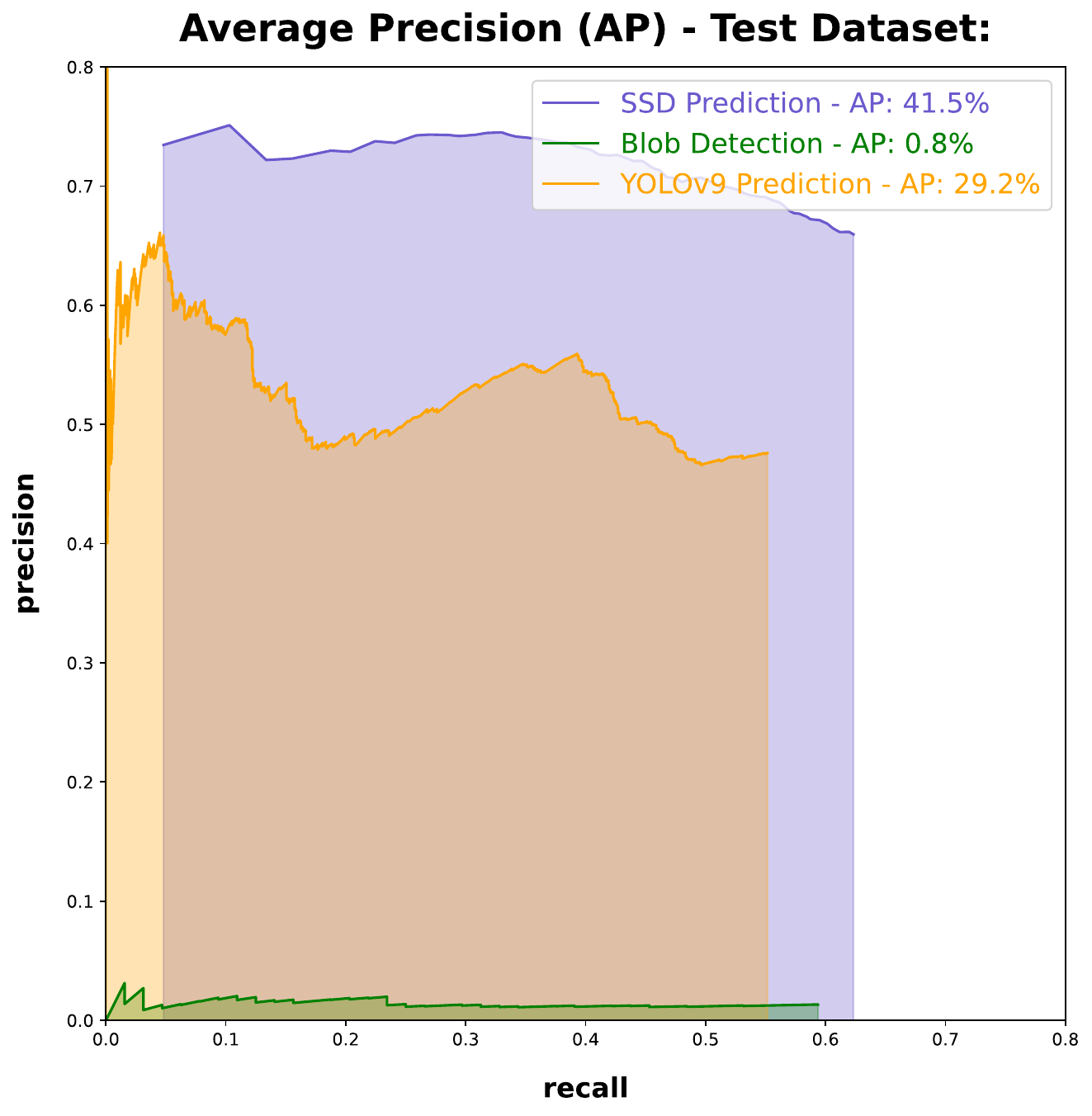}
\end{subfigure}
\caption{Comparison of our fine-tuned traditional blob detection method with the AI-based SSD300 and YOLOv9 models, using the average precision metric.}
\label{fig:mAP_comp}
\end{figure}

When comparing the classical blob detection approach with state-of-the-art AI-driven methods — here SSD300 and YOLOv9 \cite{wang2024yolov9} — a significant decrease in the false positive rate is observed, as shown in Figure~\ref{fig:mAP_comp}. Furthermore, the entire image is recognizable without any blind spots. The blob detection achieved a recall rate of 59.3\%. However, it also detected many noisy image sections as false positive objects, which significantly reduced its precision. The SSD algorithm makes individual decisions for each image and does not consider moving objects. This approach resulted in a recall rate of 62.4\% on average, which is significantly higher. This value corresponds to the highest recall achieved across all confidence thresholds, as derived from the precision-recall curve in Figure~\ref{fig:mAP_comp}.

\subsubsection{CNN classification}
For the proof-of-concept, the 680{,}000 annotated images were used for training in the first step.
The recognition rates of ML-methods for the four classes (see Figure~\ref{fig:classes}) based on independent test data show an accuracy of 98.2\%.
The misclassified ROIs are broken down as follows:
\begin{itemize}
    \item 1.1\% false positive: non-bird object recognised as a bird, and
    \item 0.7\% false negative: bird object not recognised as a bird.
\end{itemize}
The inclusion of paraglider images significantly increases the false positive rate due to their close proximity to an airfield in the vicinity of the test site and the difficulty in identifying individuals within the ROI. This presents challenges for neural network differentiation, as paragliders can be misidentified as bird silhouettes under certain viewing conditions.

\begin{figure}[H]
\centering
\begin{subfigure}[b]{0.49\textwidth}
  \centering
  \includegraphics[width=\linewidth]{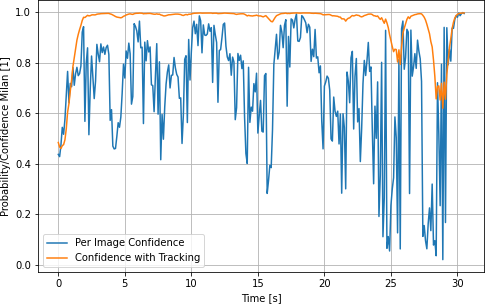}
  \caption{Recognition probability curves comparing the detection of a red kite under poor visibility conditions, with (orange) and without (blue) tracking, within the range of 200 to 700 m.}
\end{subfigure}\hfill
\begin{subfigure}[b]{0.49\textwidth}
  \centering
  \includegraphics[width=\linewidth]{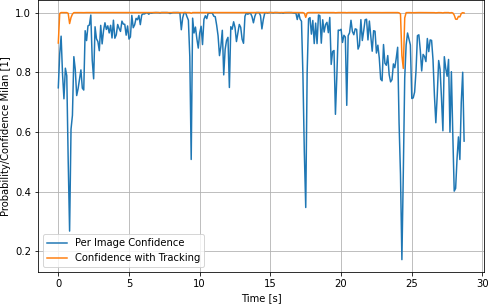}
  \caption{Recognition probability curves depicting the detection of a red kite under good visibility conditions, with (orange) and without (blue) tracking, within the range of 200 to 700 m.}
\end{subfigure}
\caption{Improving the robustness of classification by implementing a tracking method.}
\label{fig:tracking}
\end{figure}

After classification, tracking the objects increases the robustness of the prediction and subsequent decision-making for shutdown.  
To achieve this, a Kalman filter is used to track detected objects across consecutive frames. It incorporates positional and velocity information to estimate object trajectories. These trajectories enable Bayesian integration of frame-wise class probabilities, resulting in higher overall confidence and fewer misclassifications. This temporal smoothing is particularly beneficial under challenging conditions, as illustrated by the improved stability of the orange tracking curves in Figure~\ref{fig:tracking}.

\vspace{-0.7em} 

Even in challenging scenarios, the change in posture during the flight path demonstrated that a highly accurate prediction of the bird species, exceeding 99\%, was achievable within 3–4 seconds.

\section{Conclusion}
BirdRecorder represents a significant breakthrough in mitigating bird-turbine collisions, a pressing challenge in balancing renewable energy expansion and wildlife conservation. It combines high-performance AI detection (SSD and SAHI), tracking mechanisms, and hardware acceleration to detect tiny objects—particularly the red kite—up to a distance of 800\,m in real time. Our complete method surpasses both traditional and current state-of-the-art approaches in detection accuracy and efficiency.

Operational evaluations show an impressive 99\% classification accuracy after 3--4 seconds of tracking, demonstrating its effectiveness in collision prevention. To complement this validation, long-term evaluations under rare conditions such as fog or heavy rain are planned to further strengthen operational experience.

These efforts underline the broader potential of AI-based solutions to support sustainable energy development while protecting avian species, as highlighted by the success of BirdRecorder.

\newpage

\begin{credits}
\subsubsection{\ackname}
The authors thank Ursula Amann, Kay Ohnmeiss, Amit Skanda, Frank Musiol, Christian B{\"a}r, and Herbert Stark for valuable contributions, hardware support, and discussions. N.K., N.G., and F.P.G.Z. acknowledge funding from the German Federal Ministry for Economic Affairs and Climate Action for the project \textit{BBR2.0} (grant no. 03EE2047A), as well as from the Federal Agency for Nature Conservation (BfN) with resources provided by the BMUV under grant no. 3518 86 010B (\textit{BirdRecorder}). N.K. and F.P.G.Z. are supported by the state of Baden-W{\"u}rttemberg (Germany) through the project \textit{BirdRecorderValid} (grant no. L75 32102).

\end{credits}
%
%
%
%
\bibliographystyle{unsrt} 
\bibliography{literature}

\end{document}